\documentclass[letterpaper]{article} 
\usepackage[preprint]{aaai2027}
\usepackage[hyphens]{url}  
\usepackage{graphicx} 
\usepackage{natbib}  
\usepackage{amsmath}
\usepackage{amssymb}
\usepackage{caption} 
\usepackage{algorithm}
\usepackage{algorithmic}
\usepackage{xcolor}
\usepackage[most]{tcolorbox}
\usepackage[hyphens]{url}
\usepackage{graphicx}
\usepackage{natbib}
\usepackage{amsmath}
\usepackage{amssymb}
\usepackage{caption}
\usepackage{booktabs}
\usepackage{multirow}
\usepackage{placeins}

\usepackage{newfloat}
\usepackage{listings}
\DeclareCaptionStyle{ruled}{labelfont=normalfont,labelsep=colon,strut=off} 
\floatstyle{ruled}
\newfloat{listing}{tb}{lst}{}
\floatname{listing}{Listing}

\usepackage{booktabs}
\usepackage{multirow}
\usepackage{xcolor}
\usepackage{colortbl}
\usepackage{subcaption}
\definecolor{skillliftblue}{HTML}{EBF4FF}

\newcommand{\gain}[1]{{\color{gray}\tiny \textsubscript{+#1}}}

\newcommand{\base}{\color{white}\tiny \textsubscript{+00.0}}
\newcommand{\ourgain}[1]{{\color{blue}\tiny \textsubscript{+#1}}}

\newtcolorbox{promptbox}[1]{
  colback=green!5!white,
  colframe=green!60!black,
  fonttitle=\bfseries\sffamily,
  title=#1,
  enhanced,
  attach boxed title to top left={yshift=-2mm, xshift=2mm},
  boxed title style={colback=green!60!black, sharp corners},
  top=3mm,
  bottom=3mm,
  left=2mm,
  right=2mm,
  boxrule=0.5pt,
  breakable,
  toprule at break=0.5pt,
  bottomrule at break=0.5pt,
  pad before break=3mm,
  pad after break=3mm,
  before skip=10pt,
  after skip=10pt
}

\title{SkillLift: Learning Dense Rubrics from Sparse Oracles for Efficient Skill Evolution}
\author{
    Haoxiang~Kang\textsuperscript{\rm 1},
    Ming~Wen\textsuperscript{\rm 2}\corresponding
}
\affiliations{
    \textsuperscript{\rm 1}Independent Researcher\\
    \textsuperscript{\rm 2}Fudan University\\
    shaonianjingyi@gmail.com, mwen23@m.fudan.edu.cn
}

\begin{document}

\maketitle

\begin{abstract}
LLM-based agents increasingly rely on persistent skills, i.e., reusable procedural prompts, to adapt without weight updates. Existing skill self-evolution methods directly revise skill text based on execution feedback, but each oracle evaluation requires a full agent rollout, creating a supervision bottleneck that confines search to failure-patching updates. Our key insight is that ranking is a smoother supervision target than absolute outcome regression: identifying which skill is better requires fewer oracle evaluations than predicting exact scores. Building on this insight, we propose \textsc{SkillLift}, which decouples skill search from oracle cost by learning an oracle-aligned rubric as a structured evaluation space. We formalize this as a bilevel optimization problem solved via alternating optimization: an inner loop uses the frozen rubric as a cheap surrogate to guide skill revision at no oracle cost, while an outer loop invokes a small number of oracle rollouts to re-align the rubric via rank correlation, amortizing oracle cost and stabilizing text-space updates. Experiments on complex agent task benchmarks show that our method outperforms existing auto-skill methods with 40--70\% less token cost compared to frontier evolving methods.
\end{abstract}

\begin{links}
    \link{Code}{https://github.com/WalteR-MittY-pro/SkillLift}
\end{links}

\section{Introduction}

LLM-based agents have advanced rapidly in reasoning, planning, and interaction with complex environments~\cite{yao2023react,shinn2023reflexion,schick2023toolformer}. A key contributor is the use of agent skills: structured, reusable procedural prompts that package task-specific strategies, tool policies, and domain heuristics into portable artifacts~\cite{anthropic2025skills,li2026skillsbenchbenchmarkingagentskills}. A skill persists as an external adaptation layer---it is loaded before task execution, evaluated on agent rollouts, and revised over successive attempts. This persistence decouples capability from model weights, allowing a frozen agent to adapt through external text. Recent works have focused on skill self-evolution, the process of automatically improving skills from execution feedback~\cite{yang2026skillopt,lin2026muse,ma2026skillclaw}, as the core mechanism for agents to continuously self-improve.

Existing methods attempt to drive skill evolution by directly revising skill text based on execution outcomes. Approaches range from evolutionary search with surrogate verifiers~\cite{zhang2026coevo,alzubi2026evoskill}, through text-space optimizers constrained by edit budgets and validation gates~\cite{yang2026skillopt,khattab2024dspy,wang2026skillgrad}, to lifecycle-based systems that incorporate unit-test feedback~\cite{lin2026muse,liu2026skillforge}. Despite their diversity, these methods converge on a common framework: they treat skill improvement as direct empirical risk minimization over the skill document, using limited oracle feedback to guide each text revision.

\begin{figure}[t]
\centering
\includegraphics[width=\columnwidth]{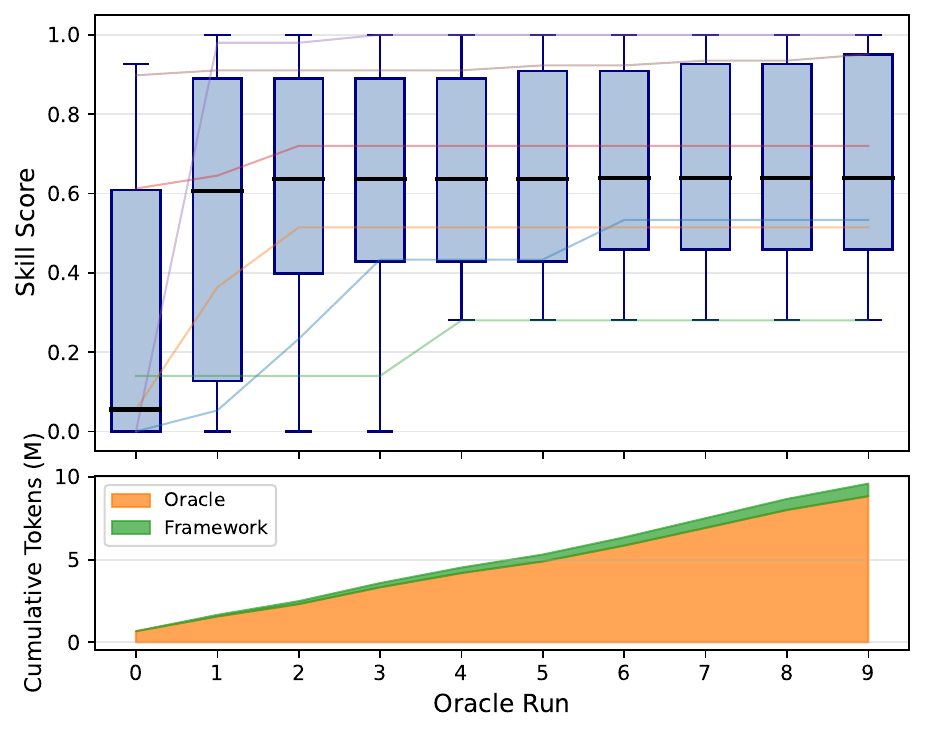}
\caption{Pilot study (10 WildClawBench~\cite{ding2026wildclawbenchbenchmarkrealworldlonghorizon} tasks, 10 rounds, 3 seeds). \textbf{Top:} per-round averaged best/mean/worst skill scores. \textbf{Bottom:} averaged cumulative oracle vs.\ framework tokens. Scores show high variance and poor scaling; oracle cost dominates and grows.}
\label{fig:pilot}
\end{figure}

This approach faces a fundamental supervision bottleneck: each oracle evaluation requires a full agent rollout, limiting the number of candidates that can be assessed~\cite{jimenez2024swebench}, while the discontinuous skill text space~\cite{yang2024opro} means few oracle signals drive conservative, failure-patching updates. We confirm this empirically with a pilot study (Figure~\ref{fig:pilot}): across 10 WildClawBench~\cite{ding2026wildclawbenchbenchmarkrealworldlonghorizon} tasks evolved for 10 rounds (3 rollout seeds per skill), skill optimization fails to scale---the best--worst score gap remains large throughout, indicating high variance---and oracle token consumption far exceeds framework overhead, worsening with each successive run. These findings motivate decoupling skill search from oracle cost via a cheap surrogate that approximates the oracle's relative preferences, allowing optimization to proceed at scale while oracle evaluations are invoked sparingly for re-alignment. Table~\ref{tab:method_comparison} contrasts \textsc{SkillLift} with representative existing methods along four dimensions: whether the method evolves skills, whether it employs a surrogate evaluator, whether it provides dense feedback, and its oracle-call complexity.

\begin{table}[t]
\centering
\caption{Only \textsc{SkillLift} combines dense feedback with amortized oracle-call complexity. $L$ = evolution steps; $T$ = evolution epochs.}
\label{tab:method_comparison}
\renewcommand{\arraystretch}{1.2}
\setlength{\tabcolsep}{3pt}
\footnotesize
\begin{tabular}{lcccp{1.2cm}}
\toprule
\textbf{Method} & \textbf{Evolv.} & \textbf{Surro.} & \textbf{Dense fb.} & \textbf{Oracle calls} \\
\midrule
Trace2Skill      & $\times$ & $\times$ & $\times$ & $O(1)$ \\
MUSE-Autoskill    & $\checkmark$ & $\times$ & $\times$ & $O(LT)$ \\
SkillOpt          & $\checkmark$ & $\times$ & $\times$ & $O(LT)$ \\
CoEvoSkills       & $\checkmark$ & $\checkmark$ & $\checkmark$ & $O(LT)$ \\
\rowcolor{skillliftblue}
\textbf{SkillLift (ours)} & $\checkmark$ & $\checkmark$ & $\checkmark$ & $\boldsymbol{O(L)}$ \\
\bottomrule
\end{tabular}
\end{table}

Our key insight is that skill evolution should not directly chase oracle outcomes; instead, it should first learn a structured evaluation space (a rubric) that captures the oracle's relative preferences among skills. Ranking is a smoother supervision target than absolute outcome regression: it requires only identifying which skill is better, not predicting its exact score, so fewer oracle evaluations suffice to learn a useful rubric~\cite{christiano2017deep}. Ranking is also robust to noisy or non-comparable oracle values, since relative order is preserved even when absolute scores are unreliable~\cite{3016100.3016209,sugiyama12_interspeech}. Once aligned, the learned rubric converts sparse oracle feedback into dense and interpretable signals for skill generation.

We formulate skill evolution as a bilevel optimization problem~\cite{colson2007overview}: the lower level optimizes a population of $K$ skills to maximize rubric-assessed quality, while the upper level aligns the rubric with oracle preferences over the resulting skill population. We solve this objective via alternating optimization~\cite{bezdek2003convergence}: in the skill-update step, the rubric is held fixed as the surrogate evaluator, providing dense feedback for improving each skill; in the rubric-update step, skills are held fixed and oracle evaluations are invoked solely to re-align the rubric-induced ranking with the oracle ranking. This decomposition amortizes oracle cost, stabilizes text-space updates, and reduces overfitting by directing skill improvements toward general quality criteria rather than instance-specific failure patches. Our main contributions are as follows. 

\begin{itemize}
  \item We identify the supervision bottleneck in direct skill optimization and propose learning an oracle-aligned rubric as a structured evaluation space to address it.
  \item We formulate skill evolution as a bilevel optimization problem and solve it via alternating updates between skill optimization guided by rubric feedback and rubric alignment guided by oracle ranking, alleviating expensive evaluations while stabilizing text-space search.
  \item We demonstrate on two complex agent task benchmarks that our method outperforms existing auto-skill self-evolving methods even when baselines are given a $2\times$ larger token budget, and achieves target performance at 40--70\% lower token cost.
\end{itemize}

\section{Related work}

\paragraph{Context optimization for LLMs.}
Treating the prompt itself as an optimization target predates the skill-setting by several years. APE~\cite{zhou2023ape} samples instruction candidates via forward LM calls and retains the best through Monte Carlo scoring; OPRO~\cite{yang2024opro} works in a similar spirit but keeps a running archive of top-scoring strings inside a meta-prompt that the LLM reads and improves at each step. DSPy~\cite{khattab2024dspy} scales the idea to full LM pipelines: it bootstraps few-shot demonstrations from successful runs and co-optimizes instruction text and example selection under a declarative signature. TextGrad~\cite{yuksekgonul2024textgrad} adds a differentiable structure to the pipeline by defining a computation graph over LM calls and backpropagating verbal critique as a surrogate gradient. All of these methods query an evaluator hundreds of times, which is feasible because each query is a single LM call. Scoring a skill candidate, by contrast, means executing the agent end-to-end~\cite{jimenez2024swebench}, and the search landscape is considerably rougher than prompt space, since a few edited words can flip a rollout from success to crash~\cite{yang2024opro}.

\begin{figure*}[t]
\centering
\includegraphics[width=\textwidth]{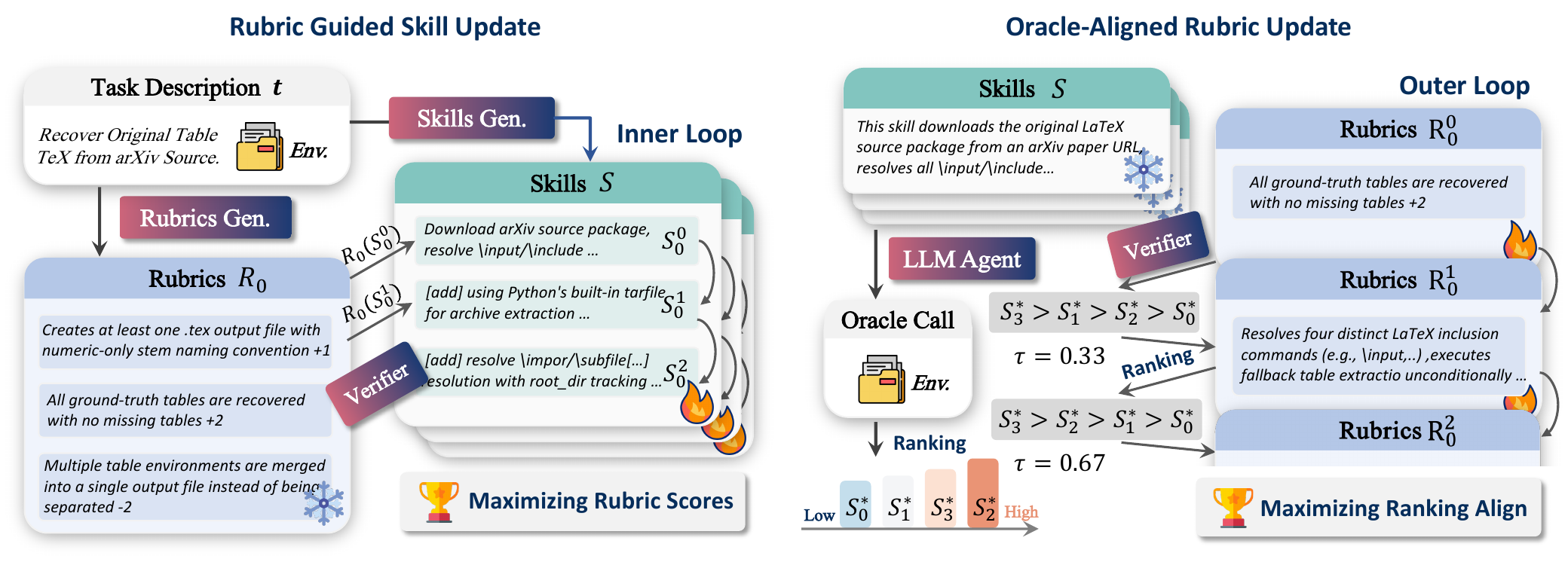}
\caption{Overview of \textsc{SkillLift}. The system alternates between two phases: \textbf{inner loop} (rubric-guided skill update) where the rubric $R$ scores and guides skill revision without oracle rollouts, and \textbf{outer loop} (oracle-aligned rubric update) where oracle evaluations on frozen skills re-align $R$ via Kendall's $\tau$.}
\label{fig:pipeline}
\end{figure*}

\paragraph{Skill extraction from trajectories.}
Several recent systems generate skills by distilling agent execution traces. Trace2Skill~\cite{ni2026trace2skill} segments a trajectory into local lessons and abstracts each into a transferable document indexed by task context. MUSE-Autoskill~\cite{lin2026muse} couples this extraction with a memory module and a skill manager that decides when to create, revise, or retire entries. SkillCAT~\cite{liu2026skillcat} attacks the attribution question head-on: for each task it pairs successful and failed trajectories, applies contrastive causal extraction to isolate the fragments responsible for the outcome gap, and routes the survivors through a topology-aware subgraph at inference. SkillFoundry~\cite{shen2026skillfoundry} sidesteps trajectories entirely, assembling skill libraries from heterogeneous scientific documents. In all of these systems, quality is judged by whether the source task passes or fails after the skill is injected---a binary verdict that costs one full oracle call per decision.

\paragraph{Text-space skill optimization.}
A more recent thread formalizes skill revision as optimization over the skill document. SkillOpt~\cite{yang2026skillopt} and SkillGrad~\cite{wang2026skillgrad} both borrow from weight-space training---the former maintains a textual learning rate, a rejected-edit buffer, and validation gates, while the latter generates text-based gradient signals from trajectory-level losses with a momentum term. SkillClaw~\cite{ma2026skillclaw} and CoEvoSkills~\cite{zhang2026coevo} shift the unit of optimization from individual skills to evolving populations, using co-evolutionary verification and cross-skill feedback to escape local optima. SkillForge~\cite{liu2026skillforge} applies iterative refinement in an industrial pipeline for cloud technical support. The shared blind spot is that every edit must be ratified by an oracle rollout before it enters the skill. With oracle calls as the binding constraint, the search collapses to conservative local patches around the current best skill. In contrast, \textsc{SkillLift} decouples skill search from oracle cost by learning a rubric that approximates oracle rankings, enabling dense feedback for skill revision without requiring a rollout per edit.

\section{Method}
\label{sec:method}

We solve the bilevel program via alternating optimization over $K$ skills and a shared rubric $R$. The initial rubric is bootstrapped from the seed skill's execution trajectory and the task description (prompt in Appendix). In the \emph{inner loop} (Mode~A), $R$ is held fixed and guides skill revision at no oracle cost. In the \emph{outer loop} (Mode~B), skills are frozen and $K$ oracle rollouts re-align $R$ with true rankings. The two modes alternate for $L$ rounds or until the champion's oracle score---the fraction of successful rollouts across seeds, which we call the \emph{pass rate}---reaches $\theta_{\text{stop}}$. Section~\ref{sec:problem} formalizes the objective; Figure~\ref{fig:pipeline} illustrates the full pipeline.

\subsection{Problem Formulation}
\label{sec:problem}

Let $\pi$ be a frozen agent and $\mathcal{T}$ a task distribution. Loading a skill $S$ before task $t$ produces a rollout $\xi \sim \pi(\cdot \mid S, t)$, scored by an oracle $\mathcal{O}(\xi)$. Each oracle call requires a full rollout, so the budget $B$ on scored $(S, t)$ pairs is the binding constraint. Directly solving $\arg\max_S \mathbb{E}_{t,\xi}[\mathcal{O}(\xi)]$ is hard: $B$ is small and the skill text space is highly non-smooth, so the optimizer sees few signals and falls back to patching observed failures.
We introduce a \emph{rubric} $R$ that scores a skill $R(S)$ via a single LLM call---no rollout needed. The rubric is cheap but must stay aligned with the oracle. We formalize this as a bilevel program~\cite{colson2007overview}:
\begin{equation}
\begin{aligned}
&\min_{R} \;\; \mathbb{E}_{t \sim \mathcal{T}} \big[ \mathcal{D}\!\big(\sigma^R_t, \,\sigma^{\mathcal{O}}_t\big) \big] \\
&\text{s.t.} \quad S^*(R) = \arg\max_{S} \; \mathbb{E}_{t \sim \mathcal{T}} \big[R(S)\big]
\end{aligned}
\end{equation}
where $\sigma^R_t$ and $\sigma^{\mathcal{O}}_t$ are the rankings that the rubric and the oracle induce over candidate skills on task $t$, and $\mathcal{D}$ is a ranking discrepancy measure (instantiated as $1 - \tau$ in the outer loop, where $\tau$ is Kendall's rank correlation). The upper level keeps $R$ aligned with the oracle's preference ordering; the lower level searches skill text to maximize $R(S)$. Because $R$ requires no rollout, thousands of candidates can be scored at near-zero cost. Only relative order matters---not score calibration---so a handful of oracle evaluations per round suffices to re-align $R$~\cite{christiano2017deep}. Once aligned, the rubric provides dense feedback for the lower-level search without consuming additional rollouts.

\subsection{Rubric-Guided Skill Update}

The lower level of the bilevel program searches skill text to maximize expected rubric score. Since $R(S)$ costs one LLM call rather than a full rollout, the skill generator can test far more candidates than the oracle budget permits. A scalar score alone, however, tells the generator \emph{how much} a skill falls short but not \emph{where}: to guide targeted revision, the evaluation must expose which aspects of the skill's behavior are deficient.

We structure the rubric as a set of $M$ binary criteria $\{(c_1, p_1), \ldots, (c_M, p_M)\}$. Each $c_i$ is a yes/no question about the skill's behavior on the task, and $p_i \in \mathbb{Z} \setminus \{0\}$ is a signed weight: $p_i > 0$ rewards meeting $c_i$ (a merit), $p_i < 0$ penalizes violating it (a flaw). Given a task--skill pair $(t, S_k)$, a \emph{verifier}---an LLM prompted with $(t, S_k, R)$---judges each criterion, producing hits $h_i(S_k, t) \in \{0,1\}$. The rubric score is
\begin{equation}
R(S_k) = \frac{\sum_{i=1}^{M} p_i \, h_i(S_k, t) - S_{\min}}{S_{\max} - S_{\min}},
\end{equation}
where $S_{\max} = \sum_{p_i > 0} p_i$ and $S_{\min} = \sum_{p_i < 0} p_i$, normalizing the score to $[0, 1]$. The hit vector $\mathbf{h}(S_k) = (h_1, \ldots, h_M)$ accompanies the score, so the generator can see which criteria failed and target those in its revision. This rubric-as-reward formulation connects to recent work on structured criteria for reward modeling~\cite{gunjal2025rubricsrewardsreinforcementlearning,liu-etal-2026-openrubrics}.

In each outer round $\ell$, SkillLift maintains a single champion skill
$S^\star_{\ell-1}$. The initial champion is the seed skill; after each Mode~B
update, the oracle selects the next champion from the refined branches together
with the preceding champion. The verifier first evaluates
$S^\star_{\ell-1}$ under the fixed rubric $R_\ell$ and produces criterion hits
$\mathbf{h}^\star_\ell$. Conditioned on this feedback, the skill generator
constructs $K$ diverse revision directions $d_{\ell,k}$---natural-language
directives targeting distinct improvement aspects (e.g., strategy optimization,
tool-usage policy, error handling), inspired by the skill lifecycle
decomposition of \citet{lin2026muse}---and derives a fresh set of candidate
branches via the generator $G$, an LLM call using the same backbone under test
as the agent itself:
\[
S_{\ell,k}^{(0)}
= G(t, S^\star_{\ell-1}, R_\ell, \mathbf{h}^\star_\ell, d_{\ell,k}),
\quad k=1,\ldots,K.
\]
Each branch is refined only within the current outer round. Let
$\mathbf{h}_{\ell,k}^{(a)}$ be the verifier hits for branch $k$ after inner
step $a$. Conditioned on these hits, the generator proposes a branch-local
revision $\widehat S_{\ell,k}^{(a+1)}$, which is retained only when its rubric
score does not decrease:
\[
S_{\ell,k}^{(a+1)} =
\begin{cases}
\widehat S_{\ell,k}^{(a+1)}
& \text{if } R_\ell(\widehat S_{\ell,k}^{(a+1)})
\geq R_\ell(S_{\ell,k}^{(a)}),\\
S_{\ell,k}^{(a)} & \text{otherwise.}
\end{cases}
\]
Thus, the verifier prevents local refinement from degrading an individual
branch, while diversity is refreshed in every outer round through new
directions. Mode~A terminates when every branch satisfies
$R_\ell(S_{\ell,k}) \geq \theta_A$ (i.e., the worst branch clears the
threshold), or when the inner-loop budget is exhausted; a per-branch early-stop
would further reduce inner-loop LLM calls but is left to future work.
Only then does Mode~B evaluate the preceding champion and the refined branches
with the oracle, align the verifier-induced ranking with the oracle ranking,
and, when necessary, revise the rubric. No oracle rollout is invoked during
Mode~A.

\subsection{Oracle-Aligned Rubric Update}

After the inner loop converges, each skill slot holds its best candidate under the current rubric. The rubric itself, however, may have drifted: it can be too lenient, too harsh, or blind to failure modes that the oracle detects. Without correction, subsequent inner-loop rounds optimize against a misaligned standard. Re-aligning the rubric requires oracle feedback, but each rollout consumes part of the oracle budget $B$.

\begin{algorithm}[tb]
\caption{SkillLift}
\label{alg:skilllift}
\textbf{Input}: task $t$, seed skill $S_0$, rubric $R$, thresholds $\theta_A, \theta_B, \theta_{\text{stop}}$, rounds $L$, population $K$, inner budget $A_{\max}$, revision budget $B_{\max}$
\begin{algorithmic}[1]
\STATE $S^\star \gets S_0$
\FOR{$\ell = 1$ \TO $L$}
  \STATE \textit{Mode A (no oracle):} score $S^\star$ with $R$; derive $K$ branches via $G(t, S^\star, R, \mathbf{h}^\star, d_k)$
  \REPEAT
    \STATE Refine each $S_k$ by $G$; accept if $R$ does not decrease
  \UNTIL{$\min_k R(S_k) \geq \theta_A$ \textbf{or} inner steps $\geq A_{\max}$}
  \STATE \textit{Mode B (oracle):} rank $\{S^\star, S_1, \ldots, S_K\}$ by oracle $\to\sigma^{\mathcal{O}}$, by rubric $\to\sigma^R$
  \WHILE{$\tau(\sigma^R, \sigma^{\mathcal{O}}) < \theta_B$ \textbf{and} revisions $< B_{\max}$}
    \STATE Revise $R$ via rubricator given $(\sigma^R, \sigma^{\mathcal{O}})$; re-score $\to\sigma^R$
  \ENDWHILE
  \STATE $S^\star \gets \arg\max_{S} \mathcal{O}(S)$;\; \textbf{break} if $\mathcal{O}(S^\star) \geq \theta_{\text{stop}}$
\ENDFOR
\RETURN $S^\star, R$
\end{algorithmic}
\end{algorithm}

\begin{table*}[t]
\centering
\caption{Score rate (\%) per task category on WildClawBench (left) and SkillsBench (right). \textbf{Bold} and \underline{underline} indicate the best and second-best per model group. \colorbox{skillliftblue}{highlighted} rows are our method. $^*$SkillsBench uses GPT-5.4-mini and WildClawBench uses GPT-5.4 to align with the official implementation.}
\label{tab:main_results}
\renewcommand{\arraystretch}{1.15}
\setlength{\tabcolsep}{2.5pt}
\scriptsize
\begin{tabular}{l *{7}{c} *{9}{c}}
\toprule
& \multicolumn{7}{c}{\textbf{WildClawBench}} & \multicolumn{9}{c}{\textbf{SkillsBench}} \\
\cmidrule(lr){2-8}\cmidrule(lr){9-17}
\multirow{2}{*}{\textbf{Method}} & Prod. & Code & Social & Search & Creative & Safety & Overall & SE & IPS & NS & OWC & Finance & Math & Cyber & Media & Overall \\
& \tiny$n$=10 & \tiny$n$=12 & \tiny$n$=6 & \tiny$n$=11 & \tiny$n$=11 & \tiny$n$=10 & \tiny$n$=60 & \tiny$n$=16 & \tiny$n$=14 & \tiny$n$=14 & \tiny$n$=14 & \tiny$n$=9 & \tiny$n$=8 & \tiny$n$=7 & \tiny$n$=5 & \tiny$n$=87 \\
\midrule

\textbf{GPT-5.4}$^*$ & 54.0\% & 47.8\% & \underline{87.6\%} & 52.3\% & 38.3\% & 38.5\% & 50.3\%\base & 26.9\% & 28.6\% & 38.3\% & 40.5\% & 18.5\% & 37.5\% & 15.0\% & 20.0\% & 29.9\%\base \\
+ Human-written & 58.8\% & 51.0\% & 85.2\% & 54.5\% & 40.4\% & 66.1\% & 56.9\%\gain{6.6} & 43.4\% & 35.7\% & 52.1\% & 47.6\% & 11.1\% & 58.3\% & 34.9\% & 40.0\% & 41.4\%\gain{11.5} \\
+ LLM-written & 56.2\% & 52.3\% & 82.5\% & 52.3\% & 40.1\% & 63.9\% & 55.7\%\gain{5.4} & 41.7\% & 33.3\% & 50.0\% & 45.2\% & 14.8\% & 54.2\% & 28.6\% & 40.0\% & 39.5\%\gain{9.6} \\
+ SkillOpt & \underline{66.0\%} & \underline{58.0\%} & 86.0\% & \underline{62.0\%} & \underline{44.0\%} & 82.0\% & \underline{64.3\%}\gain{14.0} & \underline{52.1\%} & \underline{57.1\%} & \underline{73.8\%} & \underline{69.0\%} & \textbf{33.3\%} & \textbf{66.7\%} & \underline{42.9\%} & \textbf{60.0\%} & \underline{58.2\%}\gain{28.3} \\
+ CoEvoSkills & 63.0\% & 55.0\% & 86.0\% & 57.0\% & 42.5\% & \underline{83.0\%} & 62.2\%\gain{11.9} & 50.0\% & 47.6\% & 64.3\% & 61.9\% & \underline{25.9\%} & \underline{62.5\%} & \textbf{47.6\%} & \underline{53.3\%} & 52.5\%\gain{22.6} \\
\rowcolor{skillliftblue} + \textbf{SkillLift (ours)} & \textbf{71.4\%} & \textbf{64.0\%} & \textbf{89.0\%} & \textbf{63.0\%} & \textbf{45.0\%} & \textbf{90.0\%} & \textbf{68.4\%}\ourgain{18.1} & \textbf{54.2\%} & \textbf{69.0\%} & \textbf{78.6\%} & \textbf{76.2\%} & \textbf{33.3\%} & \textbf{66.7\%} & \underline{42.9\%} & \textbf{60.0\%} & \textbf{62.5\%}\ourgain{32.6} \\

\midrule

\textbf{GLM-5.1} & 37.3\% & 52.7\% & 75.6\% & 40.9\% & 32.2\% & 62.0\% & 48.1\%\base & 39.6\% & 21.4\% & 36.8\% & 38.1\% & 14.8\% & 54.2\% & 33.2\% & 13.3\% & 32.7\%\base \\
+ Human-written & 44.9\% & 54.1\% & 78.6\% & 50.9\% & \underline{34.6\%} & 80.2\% & 55.2\%\gain{7.1} & 56.6\% & 45.2\% & 77.9\% & 52.4\% & 44.4\% & \underline{86.2\%} & 46.8\% & \underline{60.0\%} & 58.4\%\gain{25.7} \\
+ LLM-written & 41.4\% & 55.2\% & 77.5\% & 40.0\% & 33.3\% & 78.4\% & 52.2\%\gain{4.1} & 60.4\% & 42.9\% & 71.4\% & 54.8\% & 40.7\% & 83.3\% & \underline{47.6\%} & \underline{60.0\%} & 57.5\%\gain{24.8} \\
+ SkillOpt & \underline{47.5\%} & \underline{57.2\%} & \underline{83.5\%} & \underline{73.7\%} & 33.6\% & 87.9\% & \underline{62.0\%}\gain{13.9} & \textbf{66.7\%} & \textbf{64.3\%} & \underline{83.3\%} & \underline{73.8\%} & \textbf{55.6\%} & \textbf{87.5\%} & \textbf{57.1\%} & \textbf{73.3\%} & \underline{70.5\%}\gain{37.8} \\
+ CoEvoSkills & 43.6\% & 56.0\% & 82.2\% & 44.9\% & 33.9\% & \underline{88.7\%} & 55.9\%\gain{7.8} & \underline{64.6\%} & \underline{57.1\%} & 78.6\% & 66.7\% & \underline{51.9\%} & \underline{86.2\%} & \textbf{57.1\%} & \textbf{73.3\%} & 66.6\%\gain{33.9} \\
\rowcolor{skillliftblue} + \textbf{SkillLift (ours)} & \textbf{59.0\%} & \textbf{60.8\%} & \textbf{89.4\%} & \textbf{75.0\%} & \textbf{35.3\%} & \textbf{92.1\%} & \textbf{66.5\%}\ourgain{18.4} & \textbf{66.7\%} & \textbf{64.3\%} & \textbf{92.9\%} & \textbf{92.9\%} & \textbf{55.6\%} & \textbf{87.5\%} & \textbf{57.1\%} & \textbf{73.3\%} & \textbf{75.1\%}\ourgain{42.4} \\

\midrule

\textbf{DeepSeek-V4-Pro} & 43.4\% & 36.0\% & \textbf{86.8\%} & 42.3\% & 36.4\% & 41.0\% & 44.4\%\base & 40.0\% & 22.1\% & 33.6\% & 16.7\% & 18.5\% & 37.5\% & 17.4\% & 20.0\% & 26.9\%\base \\
+ Human-written & 50.6\% & 40.2\% & 83.0\% & 45.9\% & 38.7\% & 79.8\% & 53.6\%\gain{9.2} & 37.8\% & 42.9\% & 69.7\% & 50.0\% & 40.7\% & 56.5\% & \underline{56.3\%} & 53.3\% & 50.1\%\gain{23.2} \\
+ LLM-written & 44.0\% & 37.0\% & 80.0\% & 42.0\% & 38.5\% & 65.0\% & 48.3\%\gain{3.9} & 39.6\% & 40.5\% & 66.7\% & 47.6\% & 37.0\% & 54.2\% & 52.4\% & 46.7\% & 47.9\%\gain{21.0} \\
+ SkillOpt & \underline{54.6\%} & \underline{48.0\%} & 82.3\% & \underline{53.9\%} & \underline{43.8\%} & \underline{88.0\%} & \underline{59.5\%}\gain{15.1} & \underline{60.4\%} & \underline{66.7\%} & \underline{85.7\%} & \underline{78.6\%} & \underline{44.4\%} & \underline{79.2\%} & \textbf{57.1\%} & \underline{73.3\%} & \underline{69.0\%}\gain{42.1} \\
+ CoEvoSkills & 52.0\% & 40.2\% & 83.0\% & 48.0\% & 38.7\% & 83.0\% & 54.7\%\gain{10.3} & 58.3\% & 57.1\% & \underline{85.7\%} & 76.2\% & \textbf{48.1\%} & 75.0\% & \textbf{57.1\%} & 66.7\% & 66.3\%\gain{39.4} \\
\rowcolor{skillliftblue} + \textbf{SkillLift (ours)} & \textbf{63.4\%} & \textbf{51.5\%} & \underline{85.6\%} & \textbf{54.2\%} & \textbf{44.0\%} & \textbf{90.0\%} & \textbf{62.4\%}\ourgain{18.0} & \textbf{64.6\%} & \textbf{71.4\%} & \textbf{92.9\%} & \textbf{92.9\%} & \underline{44.4\%} & \textbf{83.3\%} & 52.4\% & \textbf{80.0\%} & \textbf{74.3\%}\ourgain{47.4} \\

\bottomrule
\end{tabular}
\end{table*}

The rubric need not reproduce oracle scores---it only needs to rank skills in the same order. Ranking is a weaker supervision target than score regression and is robust to the noise and non-comparability that plague absolute oracle values~\cite{christiano2017deep,3016100.3016209}. We evaluate the $K$ refined branches with the oracle; the champion's preceding oracle score is reused, so only $K$ new rollouts are needed. Together with the cached champion score, the oracle ranking $\sigma^{\mathcal{O}}$ spans $K{+}1$ skills. The verifier re-scores the same skills with the current rubric $R^{(t)}$, producing local ranking $\sigma^R$. Alignment is measured by Kendall's $\tau$:
\begin{equation}
\tau(\sigma^R, \sigma^{\mathcal{O}}) = \frac{2}{(K{+}1)K} \sum_{i<j} \operatorname{sgn}\!\big(\sigma^R_i - \sigma^R_j\big) \cdot \operatorname{sgn}\!\big(\sigma^{\mathcal{O}}_i - \sigma^{\mathcal{O}}_j\big)
\end{equation}
If $\tau \geq \theta_B$, the rubric already agrees with the oracle. Otherwise, a \emph{rubricator}---an LLM prompted to revise $R$---receives the two rankings, per-skill scores and hit vectors from both sides, and optional oracle feedback, then outputs a revised rubric $R^{(t+1)}$. The revision can add missing criteria, reweight existing ones, or remove criteria that consistently disagree with the oracle. The verifier re-scores and re-checks alignment, iterating until $\tau \geq \theta_B$ or the iteration budget is exhausted. If the budget is exhausted before alignment is reached, the last revised rubric---which maximizes $\tau$ over the iteration history---is retained; the champion is still selected directly by oracle score, so a misaligned rubric degrades inner-loop guidance quality but never corrupts the final output. With small $K$ (e.g., $K=3$ in our experiments, yielding $K{+}1=4$ ranked skills), $\tau$ takes only $\binom{K{+}1}{2}+1$ discrete values, so we set $\theta_B=0.9$ to require near-perfect agreement; the rubricator receives per-skill hit vectors as supplementary signal to compensate for the coarseness of ordinal comparison alone.

The rubricator does not see a single misjudged case. Instead, it receives a systematic discrepancy pattern: which skills the rubric over- or under-rates relative to the oracle. This pattern directly identifies the criteria responsible for the misalignment, enabling targeted revision. This iterative rubric evolution resonates with recent work on adaptive rubric refinement for reinforcement learning~\cite{shao2026drtulureinforcementlearning}.

\section{Experiments}

\subsection{Experimental Setup}

\paragraph{Benchmarks.}
We evaluate on two benchmarks differing in harness, scale, and difficulty. \textbf{WildClawBench}~\cite{ding2026wildclawbenchbenchmarkrealworldlonghorizon} provides 60 bilingual, multimodal tasks across six categories in Docker containers with a real OpenClaw CLI harness. \textbf{SkillsBench}~\cite{li2026skillsbenchbenchmarkingagentskills} contains 87 tasks across 8 domains under the OpenHands harness~\cite{wang2025openhandsopenplatformai} with deterministic verifiers, where curated skills raise the average pass rate up to 50.5\%.

\paragraph{Baselines.}
We compare against four baselines. \textbf{Human-written} skills are benchmark-provided or practitioner-authored. \textbf{LLM-written} skills are generated one-shot by GPT-5.4. \textbf{SkillOpt}~\cite{yang2026skillopt} is a text-space optimizer with bounded edits and validation gates. \textbf{CoEvoSkills}~\cite{zhang2026coevo} employs a co-evolutionary generator--verifier loop. All evolving methods share the same seed skills, namely the human-written and LLM-written variants, for a fair starting point.

\paragraph{Hyperparameters.}
SkillLift uses $L{=}3$ rounds, $K{=}3$ skills, $\leq 2$ inner steps, $\leq 2$ outer steps, $\theta_A{=}0.85$, $\theta_B{=}\theta_{\text{stop}}{=}0.9$. SkillOpt runs 2 epochs $\times$ 4 steps $\times$ 4 rollouts (1 held out), cosine learning rate 4$\to$2, turns 2/8. CoEvoSkills uses turns 1/20, oracle threshold 1.0, context cap 0.7 (200k chars), 5 final repetitions. Both baselines receive $2\times$ SkillLift's default token budget. The generator and rubricator both use the same backbone model as the agent under test, ensuring that SkillLift's gains stem from the framework rather than a stronger auxiliary model.

\paragraph{Evaluation Protocol.}
All methods use an early-stop strategy that halts when no improvement is observed for two consecutive rounds. We report mean per-task best evolved-skill scores over 3 rollout seeds. Token consumption is reported separately for oracle rollout cost and framework LLM cost.

\subsection{Main Results}

\begin{figure*}[htbp]
    \centering
    \begin{subfigure}{0.31\textwidth}
        \centering
        \includegraphics[width=\linewidth]{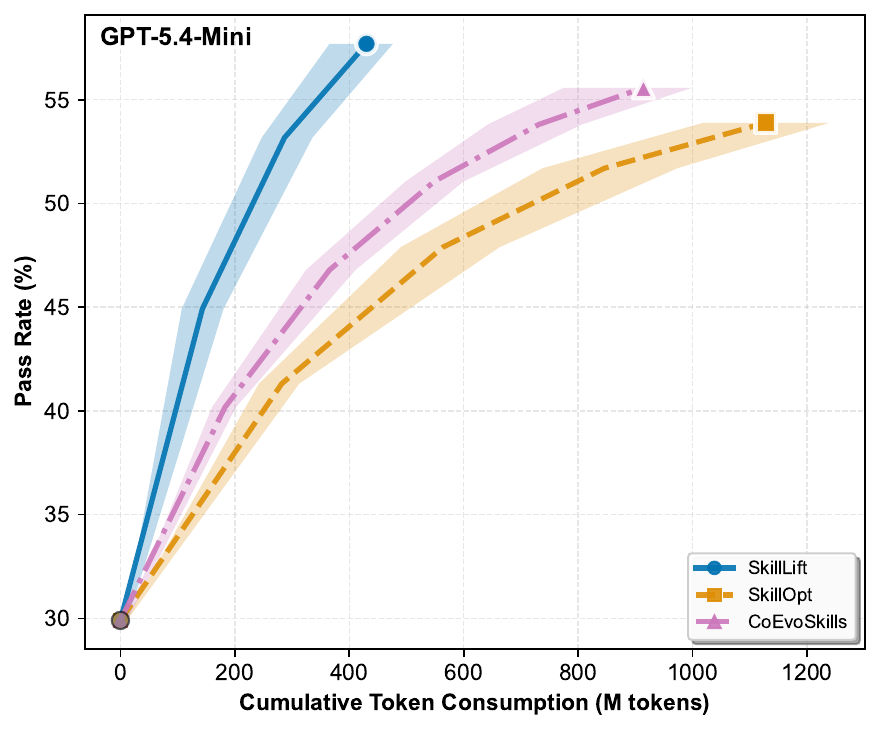}
        \caption{}
        \label{fig:token_score_gpt}
    \end{subfigure}
    \hfill
    \begin{subfigure}{0.31\textwidth}
        \centering
        \includegraphics[width=\linewidth]{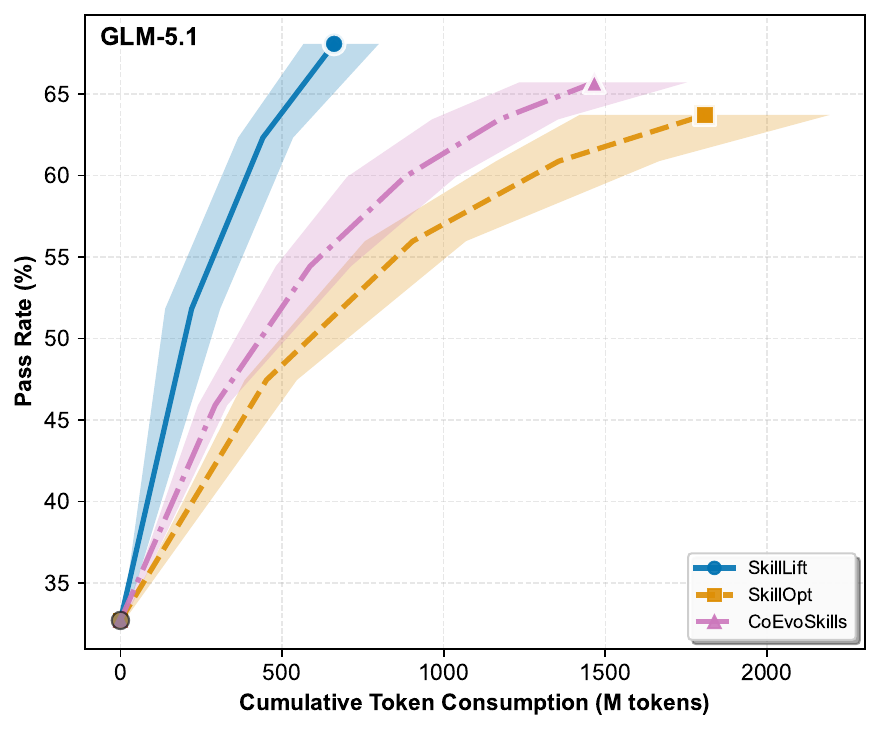}
        \caption{}
        \label{fig:token_score_glm}
    \end{subfigure}
    \hfill
    \begin{subfigure}{0.31\textwidth}
        \centering
        \includegraphics[width=\linewidth]{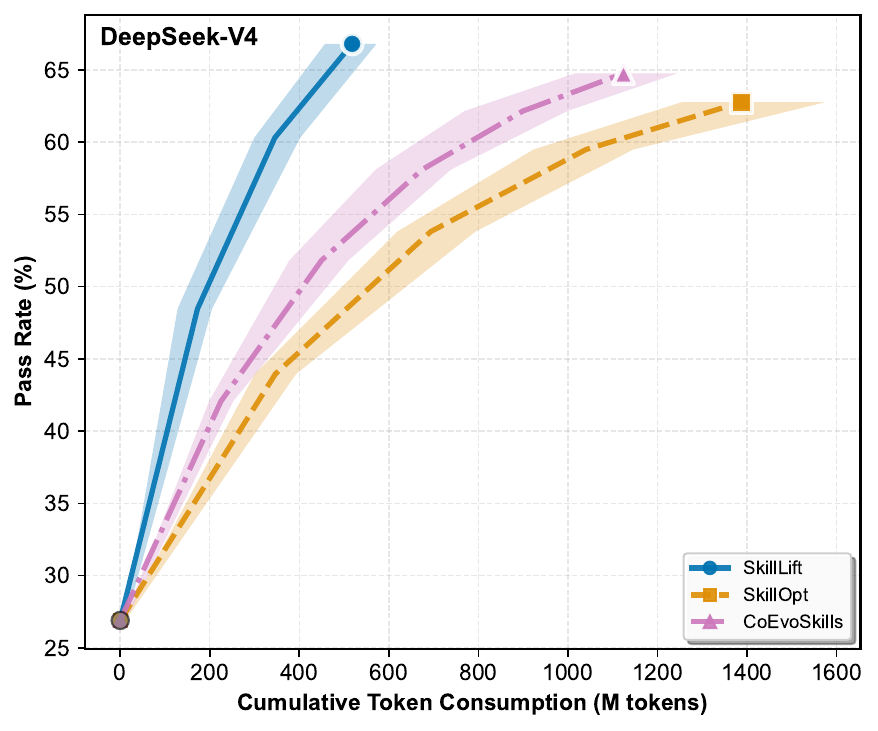}
        \caption{}
        \label{fig:token_score_dsk}
    \end{subfigure}
    \caption{Token consumption vs.\ average pass rate on SkillsBench across three backbone models. Curves are smoothed up to the 0.9 quantile and truncated when no improvement is observed for two consecutive rounds. SkillLift reaches higher pass rates at substantially lower token cost than SkillOpt and CoEvoSkills.}
    \label{fig:token_vs_score}
\end{figure*}

\begin{figure}[t]
    \centering
    \includegraphics[width=\linewidth]{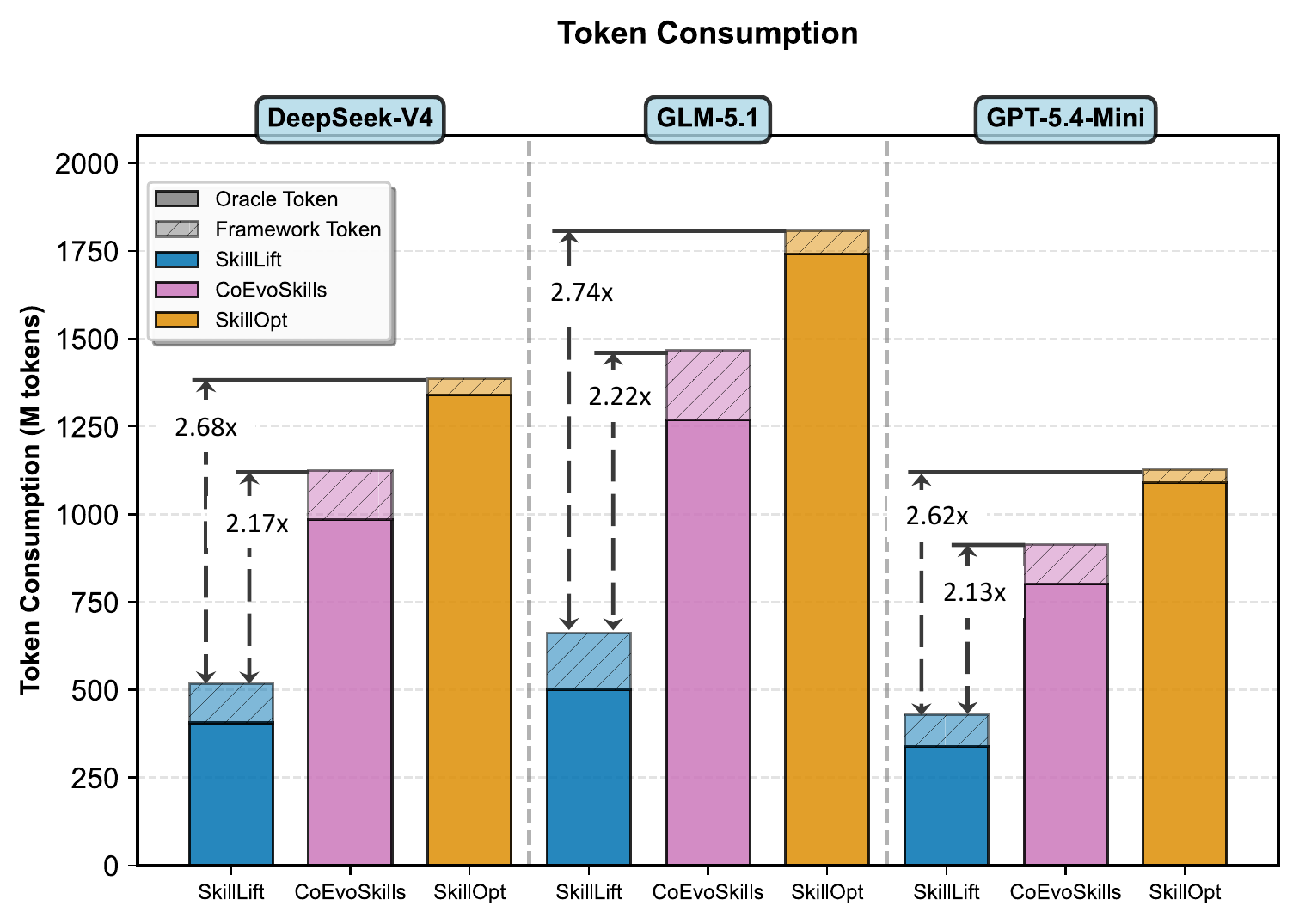}
    \caption{Average token cost to first reach the 0.9 quantile of total pass-rate score on SkillsBench. SkillLift reduces token overhead by 40--70\% compared to baselines.}
    \label{fig:token_efficiency}
\end{figure}

\begin{figure*}[h]
    \centering
    \begin{subfigure}{0.24\textwidth}
        \centering
        \includegraphics[width=\linewidth]{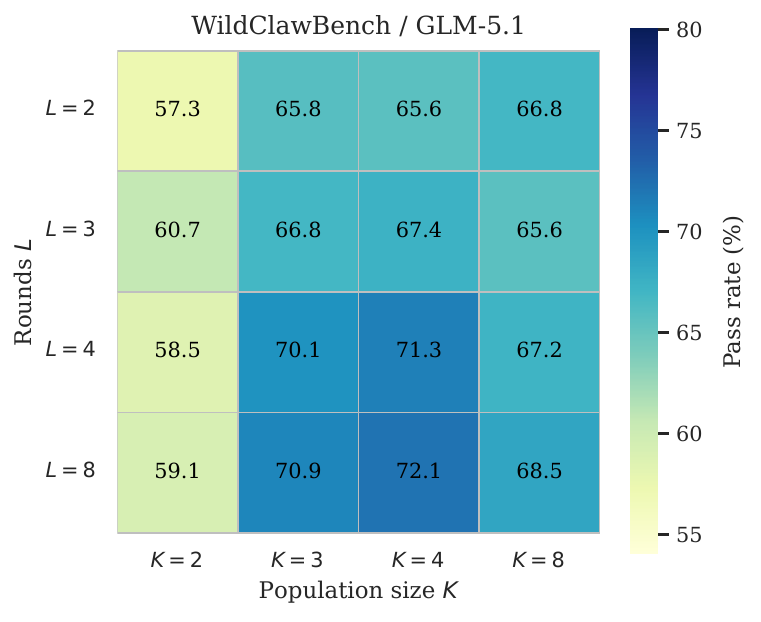}
        \caption{}
        \label{fig:heatmap_glm_wcb}
    \end{subfigure}
    \hfill
    \begin{subfigure}{0.24\textwidth}
        \centering
        \includegraphics[width=\linewidth]{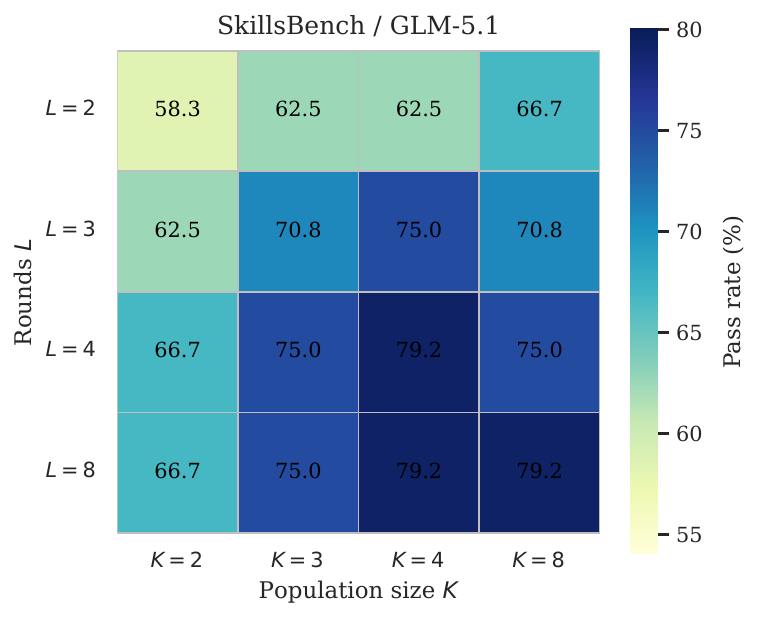}
        \caption{}
        \label{fig:heatmap_glm_sb}
    \end{subfigure}
    \hfill
    \begin{subfigure}{0.24\textwidth}
        \centering
        \includegraphics[width=\linewidth]{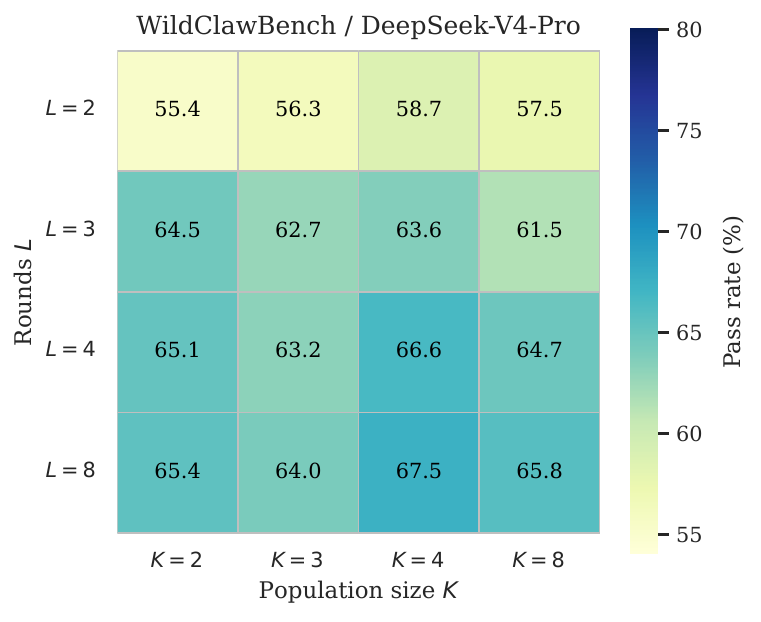}
        \caption{}
        \label{fig:heatmap_dsk_wcb}
    \end{subfigure}
    \hfill
    \begin{subfigure}{0.24\textwidth}
        \centering
        \includegraphics[width=\linewidth]{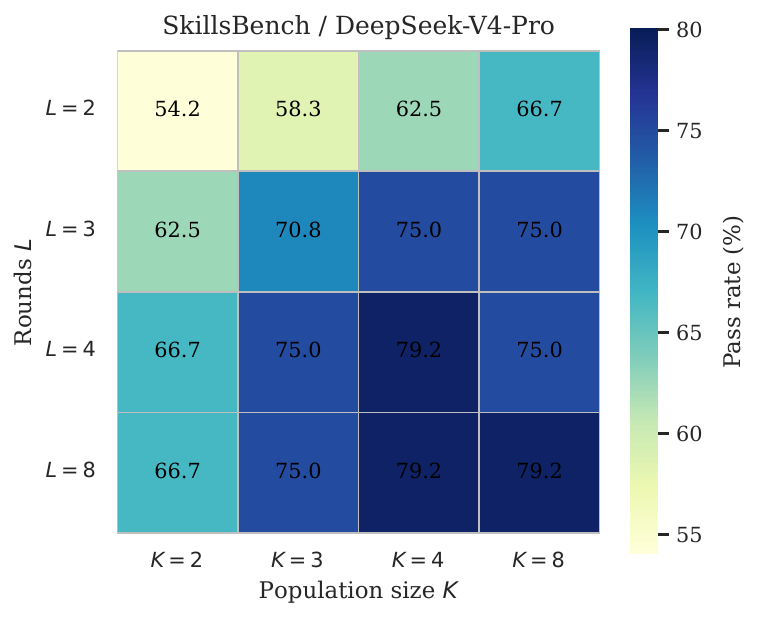}
        \caption{}
        \label{fig:heatmap_dsk_sb}
    \end{subfigure}
    \caption{Sensitivity to seed skill count and iteration rounds on SkillsBench. Seed skill counts and iteration rounds vary over \{2, 3, 4, 8\}, with overall pass rate shown for each configuration.}
    \label{fig:heatmap}
\end{figure*}

\begin{figure*}[h]
    \centering
    \includegraphics[width=\textwidth]{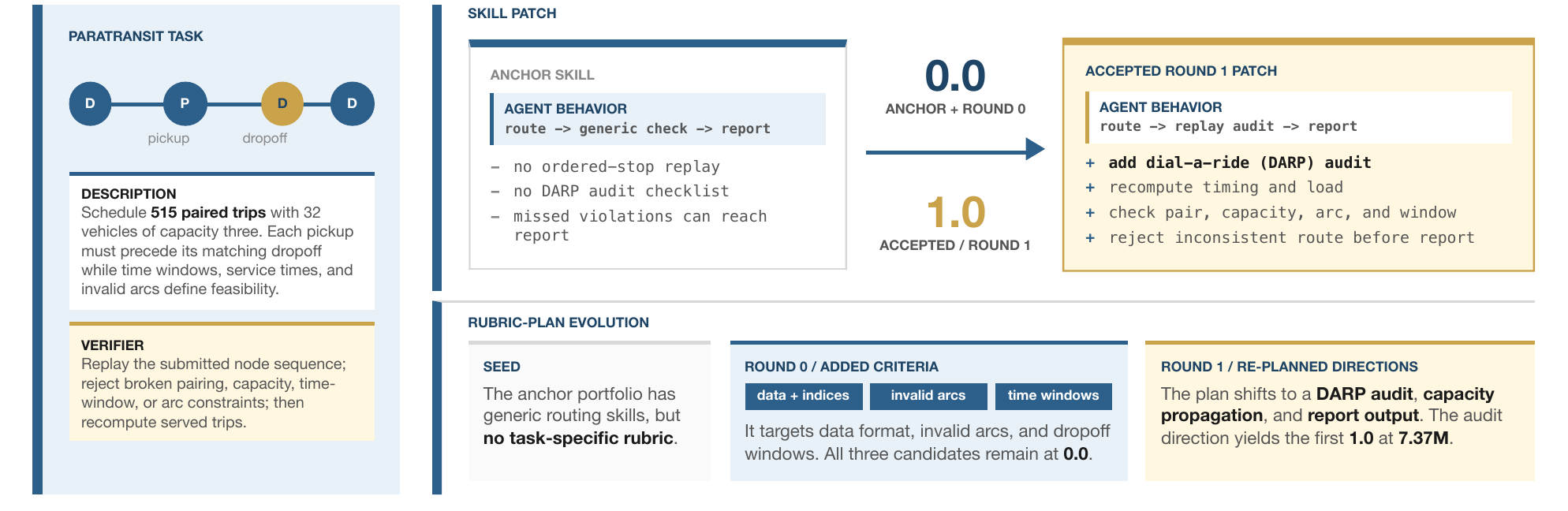}
    \caption{Case study on a SkillsBench paratransit scheduling task, showing rubric and skill co-evolution across two outer rounds.}
    \label{fig:case_study}
\end{figure*}

Table~\ref{tab:main_results} reports per-category pass rates for three backbone models across both benchmarks.

\paragraph{Static Skill Ceiling and Cross-Benchmark Gains.}
Human-written and LLM-written skills achieve nearly identical overall pass rates (differing by at most 5.3\,pp), suggesting a shared ceiling for one-shot authoring. \textsc{SkillLift} surpasses this ceiling by +8.8 to +11.5\,pp on WildClawBench and +16.7 to +24.2\,pp on SkillsBench (1.5--2.8$\times$ ratio), with larger gains on SkillsBench where deterministic verifiers provide crisper oracle signal. This confirms that the rubric's value depends on oracle quality: noisier rankings on WildClawBench's bilingual, open-ended tasks cap the achievable improvement. Crucially, \textsc{SkillLift} wins in all six model--benchmark combinations, including the strongest baseline configurations where SkillOpt and CoEvoSkills already raise pass rates substantially over static skills.

\paragraph{Heterogeneous Category-Level Gains.}
Disaggregating by category reveals pronounced heterogeneity: on GLM-5.1, Search jumps from 40.9\% to 75.0\% (+34.1) on WildClawBench and Network~Security from 36.8\% to 92.9\% (+56.1) on SkillsBench, whereas Creative barely moves (+3.1). This pattern is consistent across models---categories with code-checkable criteria see the largest jumps, while subjective Creative writing remains flat (Figure~\ref{fig:heatmap}). The rubric's binary criteria are inherently more effective when correctness is verifiable, and the rubricator naturally learns criteria that mirror the deterministic verifier. On DeepSeek-V4-Pro, the same trend holds: Code rises from 36.0\% to 51.5\% on WildClawBench, while Creative gains only 7.6\,pp despite a comparable token budget, reinforcing that subjective categories resist rubric-based improvement.

\paragraph{Convergence Speed and Token Efficiency.}
Figures~\ref{fig:token_vs_score}--\ref{fig:token_efficiency} dissect the cost-quality trade-off. \textsc{SkillLift}'s curve rises more steeply and plateaus higher than SkillOpt and CoEvoSkills, indicating that dense, criterion-level feedback guides the generator toward effective revisions early. \textsc{SkillLift} reduces token cost to reach the 0.9 quantile by 40--70\%. The saving stems from the bilevel design: rubric scoring in Mode~A costs a single LLM call per candidate, whereas baselines invoke a full oracle rollout at every step, keeping oracle-call complexity at $O(L)$ rather than $O(LT)$.

\subsection{Ablation Study}

\paragraph{Rubric Revision and Seed Quality.}
Removing the rubricator (\textit{w/o rubricator}) collapses overall pass rate to 42.5--65.5\% on SkillsBench, barely above the static-skill baseline on GPT-5.4 and GLM-5.1 (+1.1 and +1.4\,pp), confirming that oracle-aligned re-alignment---not merely dense feedback---is the core driver. The collapse is most severe on GPT-5.4, where unaligned dense feedback can actively mislead the generator when the rubric's criteria diverge from the oracle's preference ordering. Without seed skills (\textit{w/o seed skills}), performance drops 4--7\,pp: the rubricator loses its initial failure pattern and must spend early rounds discovering deficiencies that a seed already encodes. Notably, even without seeds, GLM-5.1 (71.3\%) marginally exceeds SkillOpt (70.5\%), indicating that rubric-guided revision alone can outperform scalar-score optimization.

\begin{table}[t]
\centering
\caption{Ablation study on SkillsBench (overall pass rate, \%). \textit{w/o rubricator} disables oracle-aligned rubric updates, reducing SkillLift to na\"ive self-evolution. \textit{w/o seed skills} sets the initial skill count to one with human-authored seed. \textit{w/ score regression} replaces ranking-based alignment with direct oracle score regression.}
\label{tab:ablation}
\renewcommand{\arraystretch}{1.15}
\setlength{\tabcolsep}{6pt}
\small
\begin{tabular}{lccc}
\toprule
\textbf{Variant} & \textbf{GPT-5.4}$^*$ & \textbf{GLM-5.1} & \textbf{DeepSeek} \\
\midrule
w/o rubricator & 42.5 & 59.8 & 65.5 \\
w/o seed skills & 57.5 & 71.3 & 67.8 \\
w/ score regression & 59.8 & 72.4 & 69.0 \\
\rowcolor{skillliftblue}
\textbf{SkillLift (ours)} & \textbf{62.5} & \textbf{75.1} & \textbf{74.3} \\
\bottomrule
\end{tabular}
\end{table}

\paragraph{Ranking-based Alignment.}
Replacing ranking-based alignment with direct score regression (\textit{w/ score regression}) degrades performance by 3--5\,pp despite identical oracle budget. Absolute oracle scores are noisy and non-comparable across tasks with different difficulty distributions; ranking discards the unreliable magnitude and retains only ordinal information, making it a more robust supervision target. This validates the design choice in Section~\ref{sec:method}: the outer loop aligns rubric and oracle \emph{rankings} via Kendall's $\tau$, not absolute score agreement. The two effects compound: without oracle alignment the rubric provides misaligned criteria, and without a seed the rubricator has no failure pattern to realign from, so both components are necessary.

\subsection{Case Study}

Figure~\ref{fig:case_study} traces \textsc{SkillLift} on a paratransit scheduling task from SkillsBench involving 515 paired trips, 32 vehicles, and time-window and invalid-arc constraints. The seed skill scores 0.0 because it never replays the planned stops in order and lacks a dial-a-ride audit, so capacity and window violations go undetected and propagate into the final output. In Round~0, the rubricator creates criteria for surface-level issues---data format, invalid arcs, and dropoff windows---but all candidates still score 0.0, meaning these criteria miss the root cause. After the oracle reveals that the rubric is misaligned, the rubricator replaces them with deeper checks: whether capacity is tracked across stops, whether each pair/trip respects capacity, arc, and window constraints, and whether inconsistent routes are rejected before reporting. In Round~1, the revised skill reroutes through this audit and achieves a 1.0 verifier score at 7.37M tokens.

\section{Conclusion}

We presented \textsc{SkillLift}, which decouples skill search from oracle cost by learning an oracle-aligned rubric as a structured evaluation space. A few oracle rollouts per round keep the rubric calibrated; between alignments, it scores candidates via single LLM calls with criterion-level feedback for targeted revision. On two benchmarks covering 147 tasks and three models, \textsc{SkillLift} outperforms existing auto-skill methods with 40--70\% lower token cost. Ablations confirm that oracle-aligned rubric revision is the core driver and ranking-based alignment outperforms score regression. As \textsc{SkillLift} relies on compute-time rollouts, its effectiveness is subject to model stochasticity and iteration latency may affect user experience; extending to multi-task skill libraries, integrating weight-space adaptation, and reducing rollout variance are natural next steps.

\bibliography{ref}
\appendix

This appendix gives the executable details behind SkillLift. The bilevel optimization loop holds the oracle fixed and alternates between two phases: an inner loop that optimizes skills under a frozen rubric, and an outer loop that revises the rubric to align with oracle ranking. A separate rubricator model reads verifier--oracle rank discrepancies, proposes receipt revisions grounded in visible skill evidence, and validates each candidate receipt against the frozen skill population. The oracle only receives skill execution attempts; it does not see the verifier or rubricator prompts below.

We provide the complete prompt design for each framework component, hyperparameter settings and token accounting protocol, parameter sensitivity analysis, a full case study, and per-domain result breakdowns.

\appendix

\section{Prompt Design}
\label{sec:prompts}

SkillLift orchestrates four LLM-called roles through structured prompts that enforce the bilevel separation. Table~\ref{tab:prompt-summary} summarizes their input--output contracts. All prompts return JSON under a fixed schema; the rubricator and verifier run at temperature~0 for deterministic scoring, while the generator uses temperature~0.7 for exploration.

\begin{table}[ht]
\centering
\caption{Prompt roles and their input--output contracts.}
\label{tab:prompt-summary}
\small
\begin{tabular}{l p{0.35\columnwidth} p{0.4\columnwidth}}
\toprule
Role & Input & Output \\
\midrule
Bootstrapper & task description, reference & receipt (5--15 binary criteria) \\
Verifier & skill package, receipt & criterion hits, rationale \\
Generator & skill, receipt, verifier hits & revised skill package \\
Rubricator & skills, receipt, rank gap & revised receipt \\
\bottomrule
\end{tabular}
\end{table}

\FloatBarrier

\subsection{Rubric Bootstrapper}

The bootstrapper initializes a task-specific receipt before evolution begins. Its design enforces three properties: (1)~criteria must be binary (present/absent) rather than graded, (2)~each criterion must be verifiable from the skill package alone without oracle access, and (3)~criteria must discriminate skill quality rather than restate task requirements. Signed integer points ($-5$ to $+5$, zero excluded) weight merits and flaws by importance.

\begin{promptbox}{Rubric Bootstrap System Prompt}
\small\ttfamily
You are an expert in assessment and rubric design. Generate binary, signed-point rubrics for evaluating task-specific skills. Return valid JSON only.

\smallskip\noindent
USER INPUT:
\begin{itemize}\setlength{\itemsep}{0pt}
  \item Task ID: \{task\_name\}
  \item Task Description: \{task\_description\}
  \item Reference Material: \{reference\_material\}
\end{itemize}

\smallskip\noindent
REQUIRED JSON SCHEMA:
\begin{itemize}\setlength{\itemsep}{0pt}
  \item Receipt(version, rubrics[\{rubric\_id, category, criterion, points\}], maximum\_score, minimum\_score, baseline\_score, metadata)
\end{itemize}

\smallskip\noindent
RULES:
\begin{itemize}\setlength{\itemsep}{0pt}
  \item Generate 5 to 15 binary criteria.
  \item Points: signed integers in 1..5 (merit) or -5..-1 (flaw).
  \item No fractional points or multi-threshold variants.
  \item Cover every required public task intent, including state-changing intents, mid-conversation intent changes, required preconditions, and verification obligations.
  \item Positive criteria require concrete evidence: what action, what precondition/check, when to refuse/execute, and how to verify after action.
  \item Negative criteria only for explicit unsafe or forbidden instructions present in the skill.
  \item Missing required behavior is a missed positive criterion, not a negative flaw.
  \item Do not use private oracle labels or hidden answers from task solutions.
\end{itemize}
\end{promptbox}

A repair sub-prompt handles validation failures by regenerating the receipt under stricter evidence-boundedness rules.

\subsection{Verifier}

The verifier scores a candidate skill against a frozen receipt, returning per-criterion binary hits and a short rationale. The normalized score is computed by the framework from hit weights, preventing the verifier from gaming the formula. In Mode~A (inner loop), the verifier emphasizes actionable feedback to guide revision; in Mode~B (outer loop), it applies batch-wide consistent standards to enable ranking.

\begin{promptbox}{Verifier System Prompt}
\small\ttfamily
You are an expert evaluator (role: Verifier). Evaluate the candidate skill against each rubric criterion. Return valid JSON only. Do not invent assumptions beyond the provided task, skill, and rubrics.

\smallskip\noindent
USER INPUT:
\begin{itemize}\setlength{\itemsep}{0pt}
  \item Task: \{task.to\_dict()\}
  \item Candidate Skill: \{skill package: files, entrypoint, metadata\}
  \item Receipt: \{receipt.to\_dict()\}
\end{itemize}

\smallskip\noindent
REQUIRED JSON SCHEMA:
\begin{itemize}\setlength{\itemsep}{0pt}
  \item \{"criterion\_hits": \{"r1": true, ...\}, "rationale": "short explanation"\}
\end{itemize}

\smallskip\noindent
RULES:
\begin{itemize}\setlength{\itemsep}{0pt}
  \item Positive criterion: true = merit present, false = absent.
  \item Negative criterion: true = flaw present, false = absent.
  \item Judge independently from observable package evidence only.
  \item Check for hardcoded data, keyword matching, missing tool use, and missing semantic processing.
  \item Do not credit intent without executable workflow evidence.
  \item Indeterminate positive = false; flaw not present = false.
  \item Do not assume private labels or oracle outputs.
  \item criterion\_hits keys must exactly match receipt rubric IDs.
  \item Keep rationale below 120 words.
\end{itemize}

\smallskip\noindent
MODE-SPECIFIC ADDENDUM:
\begin{itemize}\setlength{\itemsep}{0pt}
  \item Inner loop (Mode A): emphasize actionable missed positives and explicit negative flaws.
  \item Outer loop (Mode B): apply batch-wide consistent standard; do not use oracle results for scoring.
\end{itemize}
\end{promptbox}

For agent-skill formats, the anti-pattern check targets operational preconditions, observations, post-action verification, and guidance beyond procedural placeholders.

\subsection{Skill Generator}

The generator produces one candidate per population slot in each inner-loop iteration, targeting specific missed criteria identified by the verifier. It operates under two constraints that make the non-decreasing acceptance rule meaningful: edits must be substantive (no-op patches are rejected before execution), and the receipt remains frozen throughout the inner loop.

\begin{promptbox}{Skill Generator System Prompt}
\small\ttfamily
You are an expert agent skill engineer (role: Skill Generator, inner loop). You improve one candidate skill package under a fixed receipt. Return exactly one JSON object.

\smallskip\noindent
USER INPUT:
\begin{itemize}\setlength{\itemsep}{0pt}
  \item Task: \{task\}
  \item Fixed Receipt: \{receipt\}
  \item Current Skill Package: \{key, skill\_name, entrypoint, metadata, files\}
  \item Verifier Result: \{score, criterion\_hits\}
  \item Actionable Target Rubrics: \{missed positives, present flaws\}
\end{itemize}

\smallskip\noindent
OUTPUT MODE:
\begin{itemize}\setlength{\itemsep}{0pt}
  \item Choose exactly one: patch or full. Do not mix schemas.
  \item Use patch for localized edits; use full only if the package is misleading, inconsistent, or too broken to patch safely.
\end{itemize}

\smallskip\noindent
UPDATE OBJECTIVE:
\begin{itemize}\setlength{\itemsep}{0pt}
  \item Address verifier misses with executable or agent-facing changes.
  \item Do not merely rephrase rubric wording.
  \item Rescored by the same verifier and receipt.
  \item Accepted only if normalized\_score does not decrease.
\end{itemize}

\smallskip\noindent
TARGET SELECTION:
\begin{itemize}\setlength{\itemsep}{0pt}
  \item targeted\_rubrics must use exact actionable rubric IDs.
  \item Prefer executable code changes for misses involving outputs, parsing, ordering, validation, or artifact generation.
  \item Documentation-only edits allowed only for procedural misses.
  \item Metadata-only edits are invalid.
\end{itemize}

\smallskip\noindent
HARD CONSTRAINTS:
\begin{itemize}\setlength{\itemsep}{0pt}
  \item Preserve task objective, safety constraints, and working workflow unless replaced by a stronger equivalent.
  \item Do not update the receipt, hardcode outputs, use private labels, or expose hidden answers.
  \item Do not return a no-op patch.
\end{itemize}
\end{promptbox}

Population initialization uses two variants of this prompt: a seed variant grounds each initial skill in public task material, while a diversification variant assigns each of the $K$ slots a distinct workflow stress condition to prevent early population collapse.

\subsection{Rubricator}

The rubricator revises the receipt in the outer loop when verifier--oracle rank alignment falls below $\theta_B$. Unlike the bootstrapper, it receives explicit rank discrepancies (which skills the oracle separates but the verifier ties) and returns a complete replacement receipt. The central design constraint is evidence-boundedness: the rubricator may only add or strengthen criteria judgeable from visible skill text, preventing it from encoding oracle answers as evaluation rules.

\begin{promptbox}{Rubricator System Prompt}
\small\ttfamily
You are an expert rubric reviser. Improve the receipt so local skill ranking better matches public oracle evidence, without using private labels. Return valid JSON only.

\smallskip\noindent
USER INPUT:
\begin{itemize}\setlength{\itemsep}{0pt}
  \item Revision Input: \{task, current\_receipt, visible skill evidence, verifier\_scores, public oracle\_scores, verifier\_rank, oracle\_rank, rank\_alignment, rank\_mismatches, optional invalid candidate + validation report\}
\end{itemize}

\smallskip\noindent
ORACLE CONTRAST DIAGNOSIS:
\begin{itemize}\setlength{\itemsep}{0pt}
  \item If oracle scores separate skills but verifier scores tie or give full credit, the receipt is too weak.
  \item Make only evidence-bounded hypotheses from visible skill text and public scores; write no\_visible\_explanation rather than guessing.
  \item Do not infer expected answers, hidden labels, or private oracle rules from logs, artifacts, paths, or filenames.
  \item Treat tied oracle score groups as ties; never create a slot-based order.
  \item If no usable oracle contrast exists, preserve the receipt except for validation-error repair.
  \item Revise only by adding or strengthening positive binary criteria judgeable from the skill package alone.
\end{itemize}

\smallskip\noindent
REVISION RULES:
\begin{itemize}\setlength{\itemsep}{0pt}
  \item Return a complete receipt, not a patch.
  \item Keep only specific, binary-evaluable, discriminative criteria.
  \item Do not remove positive criteria; only reweight or add directly explanatory positive criteria.
  \item Negative criteria only for explicit unsafe/forbidden text.
  \item Do not invent policies, hidden rules, or unobserved execution details.
  \item Do not overfit to tools, task IDs, or a single trajectory.
  \item Repair supplied validation errors. Points remain signed integers in -5..-1 or 1..5.
\end{itemize}
\end{promptbox}

This asymmetry—adding criteria but never removing them—biases the rubric toward stricter rather than more lenient evaluation, consistent with the goal of aligning with a fixed oracle standard.

\FloatBarrier
\section{Full Hyperparameters}
\label{sec:hparams}

\subsection{SkillLift}

Table~\ref{tab:skilllift-hparams} lists all SkillLift settings. We use two Mode~B iterations so each revision is verified—a single iteration would apply an unverified final receipt.

\begin{table}[ht]
\centering
\caption{SkillLift hyperparameters.}
\label{tab:skilllift-hparams}
\small
\begin{tabular}{llr}
\toprule
Group & Parameter & Value \\
\midrule
\multirow{5}{*}{Search}
& outer rounds $L$ & 3 \\
& stable skill slots $K$ & 3 \\
& inner-loop (Mode A) iterations & 2 \\
& outer-loop (Mode B) iterations & 2 \\
& skill format & code\_package \\
\midrule
\multirow{3}{*}{Thresholds}
& Mode A acceptance $\theta_A$ & 0.85 \\
& rank alignment $\theta_B$ & 0.90 \\
& oracle-pass stop $\theta_{\mathrm{stop}}$ & 0.90 \\
\midrule
\multirow{3}{*}{Oracle}
& success threshold & 0.80 \\
& rollout concurrency & 2 \\
& feedback verbosity level & 2 \\
\midrule
\multirow{4}{*}{Evidence budget}
& budget level & auto \\
& total characters & 60{,}000 \\
& per-skill / auxiliary characters & 12{,}000 / 2{,}000 \\
& per-excerpt characters & 6{,}000 \\
\midrule
\multirow{3}{*}{Runtime}
& initial skill path & none (seeded) \\
& rate-limit retries; wait & 2; 120\,s \\
& framework role model & backbone LLM \\
\bottomrule
\end{tabular}
\end{table}

\FloatBarrier

\subsection{Baselines}

Table~\ref{tab:baseline-hparams} reports baseline settings. SkillOpt runs at reduced scale relative to its published defaults to match our per-task oracle budget.

\begin{table}[ht]
\centering
\caption{Baseline hyperparameters.}
\label{tab:baseline-hparams}
\small
\begin{tabular}{llr}
\toprule
Method & Parameter & Value \\
\midrule
\multirow{9}{*}{SkillOpt}
& epochs $\times$ steps & $2 \times 4$ \\
& rollouts per step & 4 \\
& validation rollouts per step & 1 \\
& reflection minibatch & 4 \\
& analyst workers & 2 \\
& merge batch & 8 \\
& textual learning rate & cosine $4\!\to\!2$ \\
& min / max agent turns & 2 / 8 \\
& improvement gate threshold & 0.80 \\
\midrule
\multirow{7}{*}{CoEvoSkills}
& min / max agent turns & 1 / 20 \\
& surrogate retries (max) & 15 \\
& oracle interventions (max) & 5 \\
& oracle threshold & 1.00 \\
& context cap (ratio; chars) & 0.7; 200{,}000 \\
& surrogate timeout & 30\,s \\
& final evaluation repeats & 5 \\
\bottomrule
\end{tabular}
\end{table}

\FloatBarrier

SkillOpt uses patch-style editing with a deterministic train/selection/test split at seed 42, and disables its slow-update, optimizer-meta-skill, and rejected-edit-buffer components. CoEvoSkills fixes the surrogate container image, initial-suite generation policy, and retry-counter policy.

\subsection{Token Accounting Protocol}

We count all LLM traffic uniformly: oracle rollouts plus framework calls (verifier, rubricator, reflection, surrogate verification). For the token-efficiency comparison (Figures~3--4), we measure cumulative tokens consumed before a method first reaches the $0.9$ quantile of pass-rate on a task. Tasks whose anchor already meets the threshold are excluded (no search needed); tasks that never reach it are excluded (no crossing point). Curves truncate when no improvement occurs for two consecutive rounds.

Table~\ref{tab:token-ratio} lists cost ratios. Each entry is the baseline's cumulative token cost divided by SkillLift's. Both baselines receive $2\times$ SkillLift's default budget, so the comparison is against over-provisioned baselines.

\begin{table}[ht]
\centering
\caption{Token cost to first reach the $0.9$ pass-rate quantile on SkillsBench, as a multiple of SkillLift's cost.}
\label{tab:token-ratio}
\small
\begin{tabular}{lrr}
\toprule
Backbone & SkillOpt & CoEvoSkills \\
\midrule
GPT-5.4-mini    & $2.68\times$ & $2.17\times$ \\
GLM-5.1         & $2.74\times$ & $2.22\times$ \\
DeepSeek-V4-Pro & $2.62\times$ & $2.13\times$ \\
\bottomrule
\end{tabular}
\end{table}

\FloatBarrier

\section{Extended Ablation Analysis}
\label{sec:ablation}

\subsection{Population Size and Round Count}

Table~\ref{tab:sensitivity} reports pass rate across the grid $K, L \in \{2,3,4,8\}$. Returns diminish or reverse beyond $K{=}4$: on SkillsBench/GLM at $L{=}4$, increasing $K$ from 2 to 4 gains 12.5\,pp, but 4 to 8 loses 4.2\,pp. A larger population spreads the inner-loop budget across more slots; on WildClaw/DeepSeek this spreading makes $K{=}2$ optimal at $L{=}3$. Beyond $L{=}4$, early stopping halts most tasks.

\begin{table}[ht]
\centering
\caption{Sensitivity to population size $K$ and rounds $L$ (pass rate, \%).}
\label{tab:sensitivity}
\small
\setlength{\tabcolsep}{4pt}
\begin{tabular}{lllrrrr}
\toprule
Benchmark & Backbone & $L$ & $K$=2 & $K$=3 & $K$=4 & $K$=8 \\
\midrule
\multirow{8}{*}{WildClaw}
& \multirow{4}{*}{GLM-5.1}
  & 2 & 57.3 & 65.8 & 65.6 & 66.8 \\
& & 3 & 60.7 & 66.8 & 67.4 & 65.6 \\
& & 4 & 58.5 & 70.1 & 71.3 & 67.2 \\
& & 8 & 59.1 & 70.9 & 72.1 & 68.5 \\
\cmidrule(l){2-7}
& \multirow{4}{*}{DeepSeek}
  & 2 & 55.4 & 56.3 & 58.7 & 57.5 \\
& & 3 & 64.5 & 62.7 & 63.6 & 61.5 \\
& & 4 & 65.1 & 63.2 & 66.6 & 64.7 \\
& & 8 & 65.4 & 64.0 & 67.5 & 65.8 \\
\midrule
\multirow{8}{*}{SkillsBench}
& \multirow{4}{*}{GLM-5.1}
  & 2 & 58.3 & 62.5 & 62.5 & 66.7 \\
& & 3 & 62.5 & 70.8 & 75.0 & 70.8 \\
& & 4 & 66.7 & 75.0 & 79.2 & 75.0 \\
& & 8 & 66.7 & 75.0 & 79.2 & 79.2 \\
\cmidrule(l){2-7}
& \multirow{4}{*}{DeepSeek}
  & 2 & 54.2 & 58.3 & 62.5 & 66.7 \\
& & 3 & 62.5 & 70.8 & 75.0 & 75.0 \\
& & 4 & 66.7 & 75.0 & 79.2 & 75.0 \\
& & 8 & 66.7 & 75.0 & 79.2 & 79.2 \\
\bottomrule
\end{tabular}
\end{table}

\FloatBarrier

We report $K{=}3$ rather than the grid-optimal $K{=}4$ because the 0.6--4.2\,pp gain at $L{=}3$ comes at proportional inner-loop cost. On WildClaw/DeepSeek, $K{=}2$ actually leads at $L{=}3$ (64.5 vs.\ 63.6). $K{=}3$ is a cost--quality operating point, not a tuned-per-benchmark optimum.

\subsection{Component Ablations}

Removing the rubricator collapses performance to within 1.1--1.4\,pp of the static baseline on GPT-5.4-mini and GLM. Dense feedback remains—what's missing is oracle alignment. An unaligned rubric supplies confident, wrong guidance.

Replacing ranking with score regression costs 2.7--5.3\,pp under identical oracle budget, isolating the supervision target. The effect is smallest where absolute scores are highest.

The seed ablation changes both seed quality and initial population size ($K{=}1$ vs.\ $K{=}3$), so its 3.8--6.5\,pp drop is not attributable to seed quality alone.

\FloatBarrier
\section{Per-Task Results}
\label{sec:per-task}

\begin{figure*}[p]
\centering
\begin{minipage}[t]{0.48\textwidth}
  \vspace{0pt} 
  \centering
  \includegraphics[width=\textwidth]{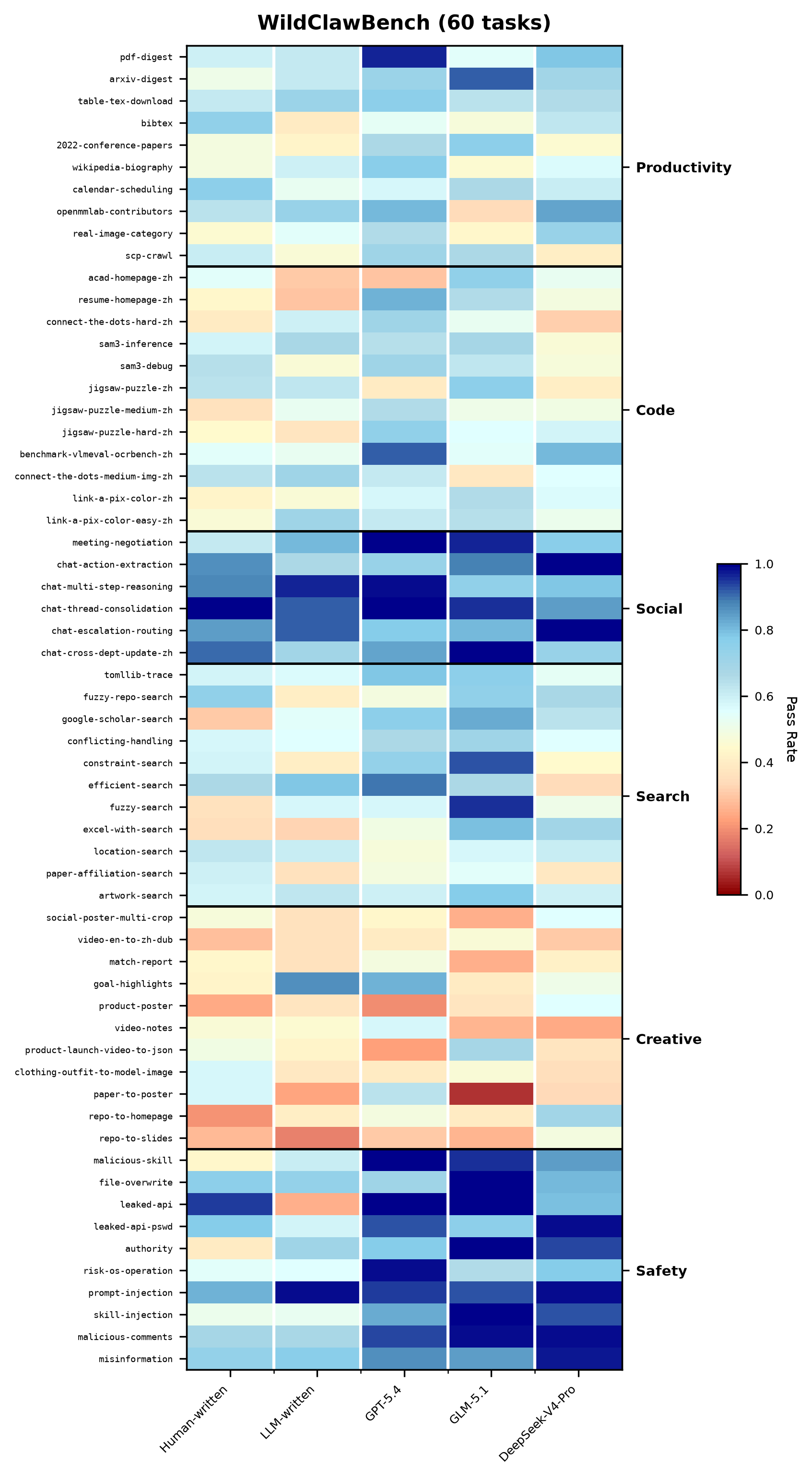}
\end{minipage}%
\hfill
\begin{minipage}[t]{0.48\textwidth}
  \vspace{0pt} 
  \centering
  \includegraphics[width=\textwidth]{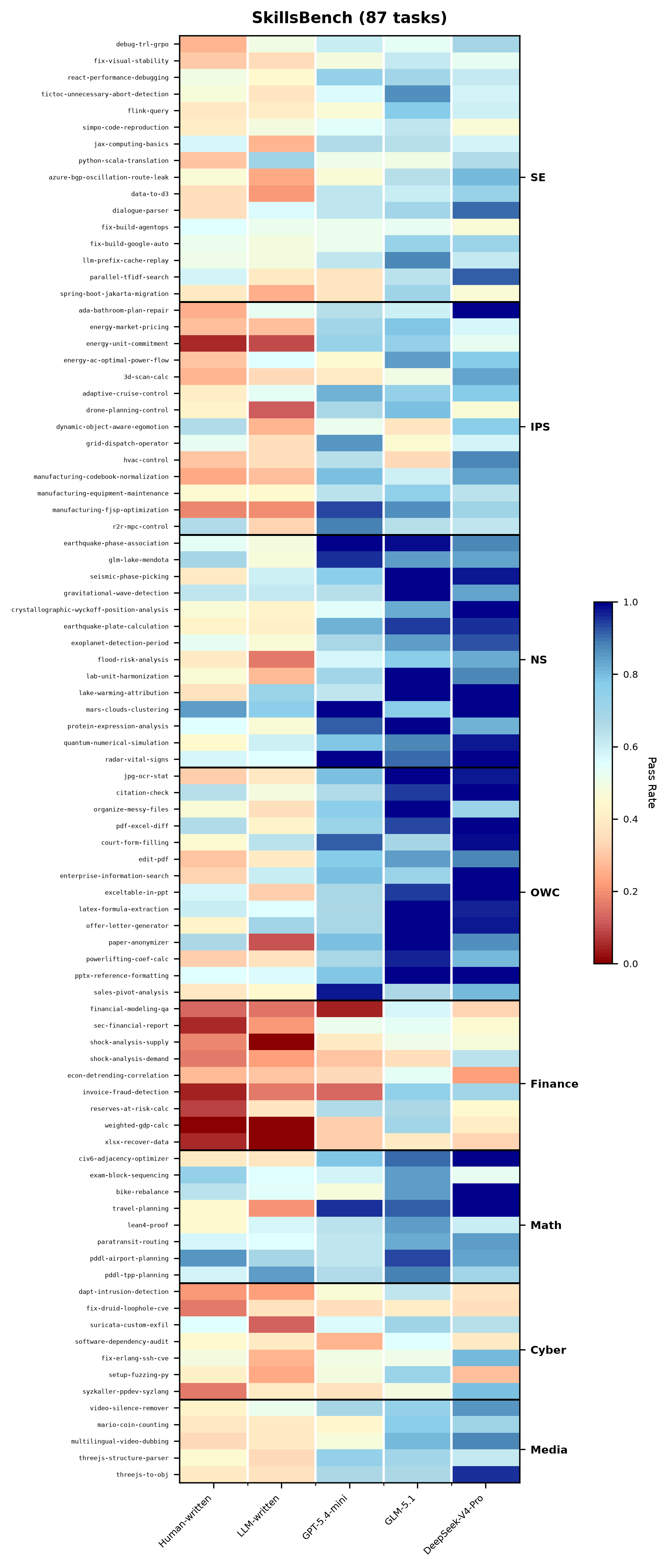}
\end{minipage}
\caption{Per-task pass rate heatmaps across both benchmarks. \textbf{Left}: WildClawBench (60 tasks across 6 domains). Columns represent \textbf{Human-written} baseline, \textbf{LLM-written} baseline, and SkillLift with three backbones: \textbf{GPT-5.4}, \textbf{GLM-5.1}, and \textbf{DeepSeek-V4-Pro}. \textbf{Right}: SkillsBench (87 tasks across 8 domains). Columns represent the same baseline conditions, with SkillLift using \textbf{GPT-5.4-mini}, \textbf{GLM-5.1}, and \textbf{DeepSeek-V4-Pro}. Task names appear on the left; domain labels on the right. Black horizontal lines separate domain boundaries. Warm colors (red/orange) indicate low pass rates ($<$0.4); cool colors (blue) indicate high pass rates ($>$0.8).}
\label{fig:per-task-heatmaps}
\end{figure*}

Figure~\ref{fig:per-task-heatmaps} shows pass rates for all individual tasks across both benchmarks. WildClawBench comprises 60 tasks spanning 6 domains; SkillsBench comprises 87 tasks spanning 8 domains. Task names appear on the left; domain labels appear on the right. Black horizontal lines separate domain boundaries. Each heatmap displays five conditions: Human-written skills, LLM-written skills, and SkillLift results across three backbones (GPT-5.4/GPT-5.4-mini, GLM-5.1, DeepSeek-V4-Pro).

\begin{table*}[htbp]
\centering
\caption{Evolution trajectory for the \texttt{tictoc-unnecessary-abort-detection} task.}
\label{tab:tictoc-trajectory}
\footnotesize
\begin{tabular}{lllrl}
\toprule
Stage & Candidate & Direction / material change & Reward & Decision \\
\midrule
Anchor & seed & Curated three-skill portfolio & 0.00 & Parent \\
\midrule
Round 0 & r000-c00 & D1: trace-driven classification. Adds TSV-to-protocol & 0.85 & Accepted \\
        &          & mapping, per-key WTS reconstruction, conservative-abort rule. & & \\
Round 0 & r000-c01 & D2: protocol-specification direction & 0.00 & Not accepted \\
Round 0 & r000-c02 & D3: safe-order witness direction & 0.35 & Not accepted \\
\midrule
Round 1 & r001-c00 & D1: implementation precision. Adds explicit data structures, & 1.00 & Terminal \\
        &          & range semantics, and edge-case handling. & & \\
Round 1 & r001-c01 & D2: access-counter ordering check & 0.85 & Not accepted \\
Round 1 & r001-c02 & D3: global serialization consistency & --- & Invalid (overlapping edits) \\
\midrule
Final 0 & frozen & Independent post-stop trial & 1.00 & Final trial \\
Final 1 & frozen & Independent post-stop trial & 1.00 & Final trial \\
Final 2 & frozen & Independent post-stop trial & 1.00 & Final trial \\
\bottomrule
\end{tabular}
\end{table*}

The heatmaps reveal several distinct patterns across task categories and backbones. SkillLift consistently outperforms both Human-written and LLM-written baselines across most domains, with particularly strong gains in procedural categories (e.g., WildClawBench Productivity, SkillsBench SE and IPS) where binary criteria effectively capture task requirements. In contrast, creative and open-ended tasks (e.g., WildClawBench Creative domain) show smaller improvements, reflecting the inherent difficulty of defining verifiable rubrics for subjective outputs.

Cross-backbone comparison shows that GPT-5.4/GPT-5.4-mini generally achieves the highest pass rates, followed by DeepSeek-V4-Pro and GLM-5.1. However, GLM-5.1 demonstrates competitive performance on specific domains such as WildClawBench Search and SkillsBench NS (network and system tasks), suggesting that domain-specific strengths persist across the SkillLift framework. The per-task granularity also exposes failure modes: tasks requiring multi-step reasoning or external tool coordination (e.g., certain Code domain tasks in WildClawBench) remain challenging for all methods, indicating that rubric-guided evolution cannot fully compensate for fundamental reasoning limitations in the underlying backbone models.

\FloatBarrier
\section{Case Study}
\label{sec:case-study}

We trace the evolution of the \texttt{tictoc-unnecessary-abort-detection} task from SkillsBench (SE domain). This task requires determining which transaction aborts are unnecessary given two trace files (write timestamps and abort records). We selected this case because its full evidence chain is preserved: anchor, search plans, candidate patches, validation records, and post-stop trials.

The receipt remains stable across both rounds (three criteria: classification procedure, trace-to-state mapping, safe-order witness). The seed portfolio scores 0.00. Round~0 candidate D1 (0.85) converts a general trace-analysis guide into a TicToc-specific method with per-key timeline reconstruction. Round~1 candidate D1 (1.00) adds explicit data structures and edge-case handling. Table~\ref{tab:tictoc-trajectory} shows the full trajectory.

The \textbf{Stage} column indicates the search round: \textit{Anchor} is the seed portfolio; \textit{Round~0} and \textit{Round~1} are inner-loop iterations with three candidate directions each; \textit{Final~0--2} are post-stop validation trials to confirm stability. The \textbf{Candidate} column identifies each skill variant (e.g., r000-c00 denotes Round~0, Candidate~0). \textbf{Direction~/ material change} describes the substantive modification applied in that candidate. \textbf{Reward} is the oracle pass rate (0.00--1.00), computed from execution rollouts. \textbf{Decision} records whether the candidate was accepted as the new parent for the next round, rejected due to insufficient improvement, marked invalid due to overlapping edits, or identified as terminal when the oracle-pass threshold ($\geq\!0.9$) was met. All three final trials achieve 1.00, confirming the converged skill's stability under independent rollouts.

The run terminates after r001-c00 meets the oracle-pass threshold. Total cost through first $\geq\!0.9$ attainment: 8.4M tokens. This case demonstrates direction-guided search and strict acceptance but not receipt revision—the initial rubric aligned without outer-loop updates.
\end{document}